\documentclass{vgtc}                          % final (conference style)

\makeatletter
\def\@fnsymbol#1{\ensuremath{\ifcase#1\or \dagger\or \ddagger\or
		\mathsection\or \mathparagraph\or \|\or \diamond\or \flat\or \ddagger\ddagger \else\@ctrerr\fi}}
\makeatother

\ifpdf%                                % if we use pdflatex
  \pdfoutput=1\relax                   % create PDFs from pdfLaTeX
  \usepackage{graphicx}                % allow us to embed graphics files
  \DeclareGraphicsExtensions{.pdf,.png,.jpg,.jpeg} % for pdflatex we expect .pdf, .png, or .jpg files
\else%                                 % else we use pure latex
  \usepackage{graphicx}                % allow us to embed graphics files
  \DeclareGraphicsExtensions{.eps}     % for pure latex we expect eps files
\fi%

\graphicspath{{figures/}{pictures/}{images/}{./}} % where to search for the images

\usepackage{microtype}                 % use micro-typography (slightly more compact, better to read)
\PassOptionsToPackage{warn}{textcomp}  % to address font issues with \textrightarrow
\usepackage{textcomp}                  % use better special symbols
\usepackage{mathptmx}                  % use matching math font
\usepackage{times}                     % we use Times as the main font
\usepackage{cite}                      % needed to automatically sort the references
\usepackage{tabu}                      % only used for the table example
\usepackage{booktabs}                  % only used for the table example
\usepackage{amsmath}
\usepackage{stfloats}

\onlineid{1182}

\vgtccategory{Research}

\vgtcinsertpkg

\title{RemoteTouch: Enhancing Immersive 3D Video Communication with Hand Touch}

\author{Yizhong Zhang\,\thanks{e-mail: yizzhan@microsoft.com} \,$^*$ \\ %
        \scriptsize Microsoft Research Asia %
\and Zhiqi Li\,\thanks{e-mail: zhiqilicg@gmail.com, work done during internship at MSRA} \,$^*$ \\ %
     \parbox{1.4in}{\scriptsize \centering Zhejiang University \\ Microsoft Research Asia} %
\and Sicheng Xu\,\thanks{e-mail: sichengxu@microsoft.com}\\ %
     \scriptsize Microsoft Research Asia %
\and Chong Li\,\thanks{e-mail: chol@microsoft.com}\\ %
     \scriptsize Microsoft Research Asia %
\and Jiaolong Yang\,\thanks{e-mail: jiaoyan@microsoft.com}\\ %
     \scriptsize Microsoft Research Asia %
\and Xin Tong\,\thanks{e-mail: xtong@microsoft.com}\\ %
     \scriptsize Microsoft Research Asia %
\and Baining Guo\,\thanks{e-mail: bainguo@microsoft.com \newline \indent \hspace{0.11111em} $^*${joint first authors}}\\ %
     \scriptsize Microsoft Research Asia
}

\teaser{
  \centering
  \includegraphics[width=\columnwidth]{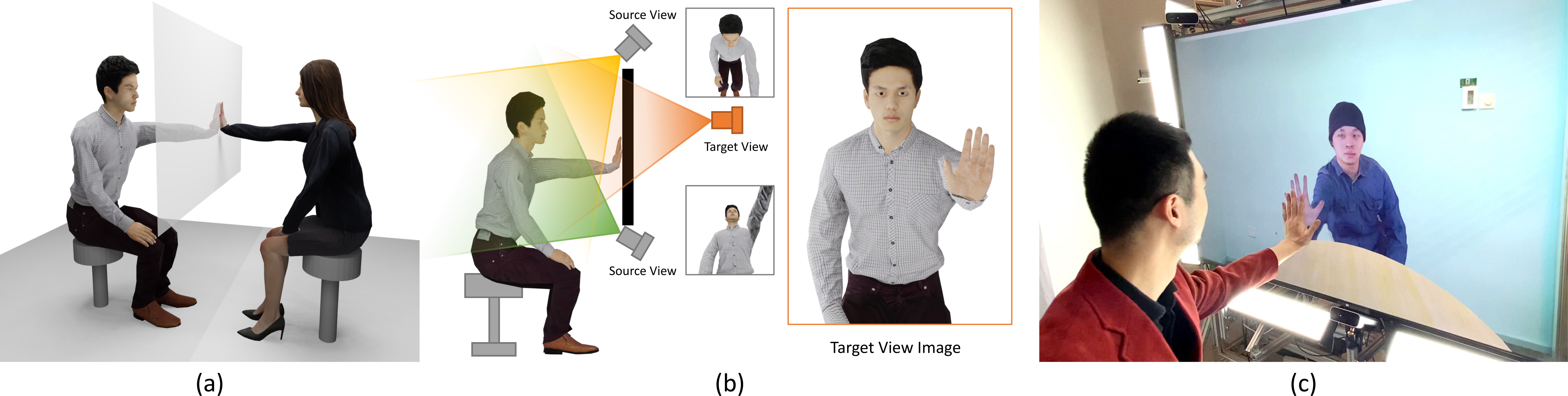}
  \vspace{-20pt}
  \caption{    
    (a) We present RemoteTouch for emulating hand touch between two remote users in an immersive 3D video communication environment as if they were sitting on different sides of a glass window. 
    (b) The multiple RGBD cameras installed around a large display provide good rendering for user's body for immersive video communication but fail to capture and render the hand as it closes to screen. 
    (c) We propose dual hand representation for rendering realistic hand as it touches the screen. The remote touch between two remote users is tested in our prototype system. 
}
  \label{fig:teaser}
}

\abstract{Recent research advance has significantly improved the visual realism of immersive 3D video communication. In this work we present a method to further enhance this immersive experience by adding the hand touch capability (“remote hand clapping”). In our system, each meeting participant sits in front of a large screen with haptic feedback. The local participant can reach his hand out to the screen and perform hand clapping with the remote participant as if the two participants were only separated by a virtual glass. A key challenge in emulating the remote hand touch is the realistic rendering of the participant’s hand and arm as the hand touches the screen. When the hand is very close to the screen, the RGBD data required for realistic rendering is no longer available. To tackle this challenge, we present a dual representation of the user’s hand. Our dual representation not only preserves the high-quality rendering usually found in recent image-based rendering systems but also allows the hand to reach to the screen. This is possible because the dual representation includes both an image-based model and a 3D geometry-based model, with the latter driven by a hand skeleton tracked by a side view camera. In addition, the dual representation provides a distance-based fusion of the image-based and 3D geometry-based models as the hand moves closer to the screen. The result is that the image-based and 3D geometry-based models mutually enhance each other, leading to realistic and seamless rendering. Our experiments demonstrate that our method provides consistent hand contact experience between remote users and improves the immersive experience of 3D video communication.
} % end of abstract

\CCScatlist{
  \CCScatTwelve{Human-centered computing}{Collaborative and social computing}{}{};
  \CCScatTwelve{Computing methodologies}{Computer graphics}{Graphics systems and interfaces}{Virtual reality}
}

\begin{document}


%\firstsection{Introduction}

\maketitle

\section{Introduction}

The ultimate goal of a teleportation system is to enable people in different locations to meet with each other as if they were in the same room. Studies \cite{Buxton1997,mehrabian2017nonverbal} have shown that to achieve this goal, the system should faithfully capture facial expressions and body gestures and support natural eye contact between remote meeting participants. Recent advances \cite{Lawrence21,Virtualcube2022} in 3D telepresence and 3D video communication make it possible to create convincing visual illusion that remote meeting participants are in the same room. While these advances are encouraging, much work remains. In particular, existing methods still cannot support physical touch between remote participants, which is an important form of nonverbal communication for real-world social interactions \cite{sin2013human,haans2006mediated,GALLACE2010246}.

In this paper, we present RemotelTouch, a method for emulating hand touch 
between two remote participants in an immersive 3D video communication environment. As shown in Fig.\ref{fig:teaser}, our method renders the life-size remote participant on a large display according to the local user’s view as if the two users were sitting on two different sides of a virtual glass window. As the remote user raises her hand to touch the screen, the local user can perceive her action and respond by raising his own hand accordingly. The virtual hand touch happens when both users’ hands contact the screen and their hands overlap on the virtual glass window. The resulting touch sensation is rendered as a screen vibration similar to the haptic rendering of iPhone touch, although other renderings of the touch feedback are certainly possible.

A key challenge in emulating the hand touch between remote participants is the realistic rendering of the participants’ hands and upper bodies as the hands approach and touch the screen. For a believable touch experience, the rendering needs to be both realistic and seamless as the hand touch the screen. This rendering task is highly non-trivial. Existing image-based rendering solutions such as \cite{Lawrence21,Virtualcube2022} can realistically render the participants hands when they are positioned away from the screen (above a distance threshold) and thus visible by the RGBD cameras that provide the color and depth video feeds to the rendering algorithms. However, these rendering solutions stop working when the hand is too close to the screen (below the distance threshold and thus invisible to the RGBD cameras). Although 3D model based solutions such as \cite{Wood2016} can successfully track the hand motion as the hand reaches the screen, the rendering quality is unsatisfactory and inferior compared to image-based rendering solutions.

To tackle this challenge we introduce a dual representation of the hand consisting of three components: an image-based model, a 3D geometry-based model, and a distance-based model fusion scheme that seamlessly combines the image- and geometry-based models. The first component, the image-based model, is a Lumigraph of the hand constructed using image and geometry data obtained by a set of RGBD cameras. Similar to \cite{Virtualcube2022}, we use a sparse set of RGBD cameras installed around the screen to capture the image and geometry data needed by the Lumigraph construction. The second component of the our dual representation, the 3D geometry-based model, is obtained by tracking the movement of the hand skeleton (including the arm and fingers) with a sideview camera and rigging the resulting skeleton. Finally, the third component of the dual representation, the distance-based model fusion, ensures the seamless fusion of the image- and geometry-based models. Specifically, when the hand is away from the screen, the image-based model is used for rendering. As the hand moves closer to the screen, the geometry-based model is given increasingly more weights in rendering and when the hand is completely invisible from the RGBD cameras, the geometry model gains the full weight. The model fusion scheme is also responsible for fitting the geometry and texture of the geometry-based model to the image-based model in the initial phase of the hand touch operation, when the hand is still away from the screen and hence fully visible from the RGBD cameras.

We implement our method and test the virtual hand touch between participants sitting at two locations. Our studies demonstrate that our design provide immersive communication experience. The virtual hand touch enabled by our method efficiency enhances the nonverbal communication of the users.

\section{Related Work}

Our work is related to multiple research fields such as 3D video communication, VR/AR, and human-computer interaction. In this section, we briefly review the techniques and systems that are closely related to our work. For comprehensive surveys of the advances in related fields, readers are referred to \cite{HapticVR2021,HapticAR2021,RemoteCollabSurvery2021,gallace2022social}.
    
\subsection{Immersive 3D Video Communication}
Many techniques and systems have been proposed for immersive teleconferencing between the participants at different locations. Early methods \cite{baker2002coliseum,gibbs1999teleport,sadagic2001tele} enable 3D video communication by capturing the geometry and appearance of the participants and rendering them into one shared virtual environments. Later, a number of approaches have been introduced to enhance the gaze contacts between the participants in various meeting setups \cite{nguyen2005multiview,kuechler2006,Jones2009,Pluss2016,kuster2012,gotsch2018telehuman2,wen2000,Towles2002,kauff2002immersive,Zhang2013} or enhance the rendering quality of human characters \cite{collet2015high, maimone2011encumbrance, zollhofer2014real, Orts2016, dou2016fusion4d, Guo2017}. Most recently, Lawrence et al. presented Project Starline \cite{Lawrence21} for high fidelity 3D communication between two participants at different locations with the help of dedicated hardware setup for 3D video capturing and light field display. Zhang et al. introduced VirtualCube \cite{Virtualcube2022} that supports immersive 3D video communication between remote participants in various meeting setups based on off-the-shelf devices. Although these techniques and systems efficiently improve the nonverbal communications between the participants with natural gaze contact and high fidelity facial expression, none of them supports social touch between the users. Besides, their setups cannot capture human hands as they move close to screen. 

Different from existing 3D video communication solutions, our method focuses on emulating hand clapping between remote users in immersive 3D video communication. We present a dual representation for modeling and rendering user's hand to support visually and physically consistent hand touch experience.  

\subsection{Hand Rendering in Remote Collaboration}

Existing methods of remote collaboration with hands can be categorized into two categories: those designed for the shared tabletop scenario, and those for the shared whiteboard scenario.
For the shared tabletop scenario, many approaches \cite{Ishii1991,Tang2007,Coldefy2007,Izadi2007,Sasa2012,Genest2013,Leithinger2014,Onishi2017SpatialCA, Iwai2018} have been proposed for capturing and visualizing the hand gestures, in which the hand embodiment captured by the cameras placed above the table is rendered over the remote table via different display devices (e.g., projector, screen, or other devices \cite{Onishi2017SpatialCA,Leithinger2014}). Le et al. \cite{Le2017} developed a mobile setup by capturing the user's hand interactions above a tablet screen via the built-in front facing camera and a mirror. Several methods \cite{Ishii1991,Coldefy2007,Izadi2007,Leithinger2014, Iwai2018} apply another vertical screens to display the remote participants' body and face to enhance the user experience. Leithinger et al. \cite{Leithinger2014} developed a physical 3D display for rendering 3D arm embodiment and objects captured from the remote side. Iwai et al. \cite{Iwai2018} further improved the user experience by rendering the upper limb image of the remote participants according to its relative position to the shared document on the local desktop and extending the rendered limb image from the vertical image of the remote participant. All these methods assume that all participants' views are on the same side of their upper limbs. As a result, they cannot be used in our scenario where the local participant's view is on the opposite side of the remote participant.

In the shared whiteboard scenario, a large display at each site is typically used as a semi-transparent whiteboard for local users to input the contents and for rendering the remote collaborators standing at the opposite side of the whiteboard. Early methods \cite{Ishii1992,Tang91,Tan2009} developed for this scenario place the camera behind a semi-transparent screen to directly capture the hand gesture, face and body of the remote participant from local user's view. Although these methods demonstrate convincing 3D remote communication and collaboration capabilities, their device setup requires larger spaces and specialized screens. Later, a few methods place the cameras in front of the screen for capturing the remote participants. 
For example, the 3D-Board system \cite{Zillner2014} uses two RGBD cameras installed on the top corners of the large display to capture the 3D geometry and texture of the remote participant and render it from local user's view. However, the cameras fail to capture the user's hand when it is close to or touches the display. Higuchi et al.\cite{Keita2015} use a single RGBD camera mounted on the side of a large display for capturing the participant. Unfortunately, the noisy and incomplete 3D model captured by the RGBD camera fails to realistically reproduce the appearance of the remote participant from local user's view. Wood et al. \cite{Wood2016} tracked the 3D hand motion of a remote user via a depth camera and then render a non-photorealistic 3D hand model driven by the tracked motion over the shared workspace.

Similar to the shared whiteboard scenario, our work aims to emulate the hand clapping of two participants sitting at the opposite side of a virtual glass. However, different from all these methods that model the hand either by image-based representation or geometry-based 3D model and are designed for offering accurate pointing position or hand pose, our method tries to deliver a visually and physically consistent touch experience. A dual hand representation and rendering scheme is proposed for realistic hand rendering.

\subsection{Social Touch in VR and AR}

A number of works \cite{alsamarei_2021,seinfeld2022evoking,Erp2015,Price2020,Sykownik2020} investigate the impact of social touch in VR and AR and validate that the social touch can enhance the user experience in remote communication and collaborations.  
Other works develop techniques for emulating social touch between remote users \cite{nakanishi2014remote, bevan2015shaking, oh2016hand}. Nakani et al. \cite{nakanishi2014remote} used a robot hand to emulate hand shaking between remote users and proved that hand touch improves the experience of video communication. Bevan et al. \cite{bevan2015shaking} set a humanoid robot as the avatar of a remote user for emulating the hand shaking between the remote users. Oh et  al.~\cite{oh2016hand} presented avatar animation techniques to support handshaking between remote users in a shared VR environment.  
Gallace et al. \cite{gallace2022social} presented a survey of the devices and studies of social touch in VR applications. Different from these methods that only focus on emulating tactile feedback of hand touch between remote users, our work aims to provide hand touch with consistent visual and haptic experience in immersive 3D video communication.  

\subsection{3D Hand Tracking}

As a critical technique for animating and rendering the synthetic hand in virtual reality and remote collaboration, vision-based 3D hand tracking has been extensively studied in computer vision field in the past decades \cite{huang2021survey}. A set of real-time methods \cite{wang2020rgb2hands,tang2021towards} have been developed for tracking 3D hand poses from monocular video at the cost of pose accuracy. Other methods \cite{khaleghi2022multi,lim2020camera} improve the robustness and accuracy of real-time 3D hand tracking with the help of multiple-view video input. However, the specialized multiple view setup limits their usage in many applications. Smith et al. \cite{10.1145/3414685.3417768} introduced a multi-view method for high accurate 3D hand tracking. However, their method cannot be performed in real time.

In our work, we selected an off-the-shelf solution provided by LeapMotion \cite{leapmotion} to track 3D hand pose of the user due to the simplicity of the device setup and software development. Since 3D hand tracking is not the contribution of our work, any other real-time 3D hand tracking methods can be used in our method.

\subsection{Haptic Rendering}
Various force feedback devices have been developed for rendering haptic in human-computer interaction tasks, including special-purpose haptic controllers \cite{Burdea1992,Massie1994ThePH,Lindeman2004,Benko2005}, in-air haptic \cite{Sodhi2013,Carter2013}, multi-sensory wristband \cite{Pezent2019}, and physical geometry primitives \cite{Cheng2017}. The goal of our work is not to develop new force feedback devices or haptic rendering methods. We thus leverage existing vibration solutions to provide the haptic feedback in our solution. 

Several approaches have been developed for detecting the tactile interactions between the user's hand and physical surfaces \cite{Kim2014RetroDepth3S, MRTouch2018} or capacitive touchscreens \cite{Streli2021}, which are then used as the input for different applications \cite{MRTouch2018,Ahuja2021,Schmitz2021}. In this work, we apply off-the-shelf hardware (e.g., a Leap Motion camera) to track 3D hand motion and detect the contact between hand and screen.

\section{Dual Representation for Emulating Touch}
Our work in this paper extends existing immersive video communication systems by providing emulated hand touch. For immersive experience, our system equips with a large display in front of each user, as shown in Fig.~\ref{fig:teaser} (c). The display has dual purposes: it not only displays the life-size 3D portrait video of the remote user with respect to local user's viewpoint but also serves as the physical media for hand contact. A set of RGBD cameras are installed around the display to capture the appearance and rough geometry of the user for rendering the 3D video of the user, and a side camera is used to track hand motion. Although our method simplifies the surrounding three-display setup in \cite{Virtualcube2022} to one frontal display, the 3D rendering of the remote user displayed on the screen will follow the local user's viewpoint and still provides immersive experience for tele-communication.

\begin{figure}[t]
    \centering
    \includegraphics[width=\columnwidth]{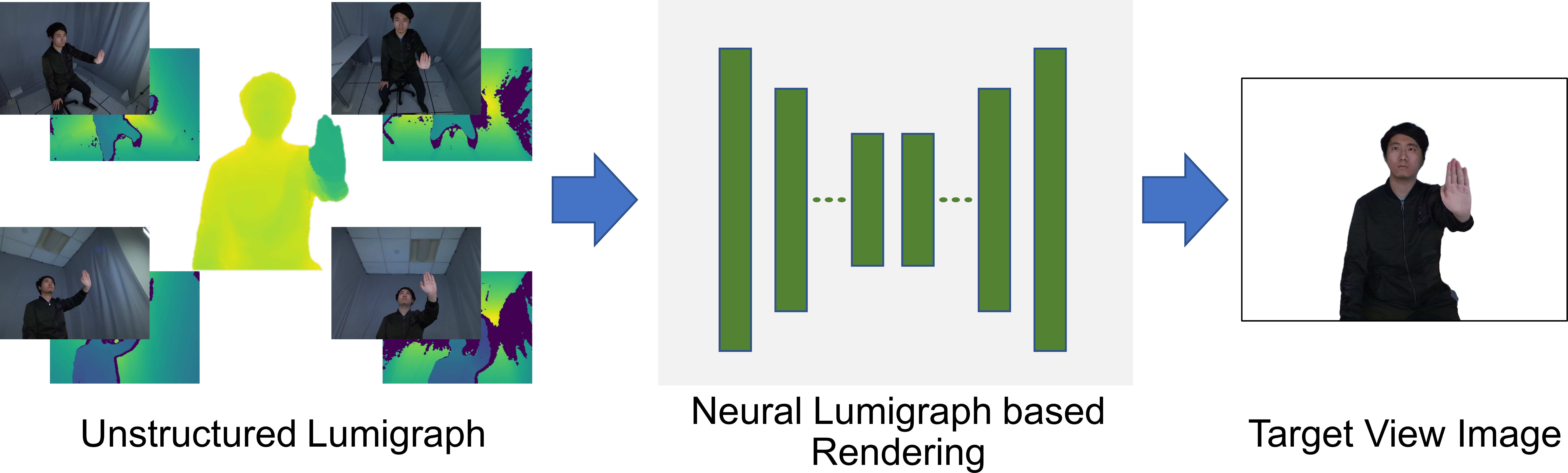}
    \caption{
    The body and hand are modeled by an unstructured lumigraph model that consists of a geometry proxy and multiple RGB images (left). The geometry proxy is aligned with the target view and refined from the input RGBD images. A neural lumigraph-based rendering algorithm (middle) is used to render the lumigraph to RGB$\alpha$ image at the target view (right). 
    }
    \label{fig:view_synthesis}
\end{figure}

A key challenge for hand touch emulation is high quality rendering of the users' hands and upper body. A realistic and temporally-smooth hand rendering throughout the virtual touch process is of paramount importance to maintain the immersive experience. To achieve this goal, we propose a dual hand representation which consists of an image-based model that uses a set of multiview images for rendering and a 3D geometry model that provides mesh rigging and rendering. A distance based fusion algorithm is applied to achieve seamless fusion of the renderings from the two models of our dual representation.

\subsection{Dual Representation of Hand}
\label{sec:handrendering}

A touch process can be roughly divided into three phases: the starting phase where the users begin to raise their hands, the approaching phase where the users move the hands towards the display, and the ending phase where the palms touch the screens and physical feedback is triggered (if real and rendered hands coincide). 

Our dual hand representation has two rendering models, one image-based and another 3D geometry-based. The former takes the multiview RGBD images captured by the cameras installed around the display as input and synthesize the target-view hand image. It provides high-quality, photorealistic results when the hand is well within the shared field of view of all RGBD cameras (e.g., in the starting phase). The latter uses a predefined hand mesh model driven by the real hand's motion for rendering, which is a realistic substitute when the hand is completely invisible to the cameras (e.g., in the ending phase). To seamlessly fuse these two models and their renderings, we adapt the texture of the mesh model to the real hand's appearance and develop a distance-based fusion algorithm to smoothly transit the former to the latter when the hands get closer to the displays (e.g., in the approaching phase).

\subsubsection{Image-based Model}
\label{sec:imagehand}

When hands and other parts of the human body are visible to the RGBD cameras (we assume the latter to be always visible), an image-based model produces the best rendering result due to the model’s capability to support photorealistic image synthesis and handle view-dependent texture appearance commonly seen in non-Lambertian surfaces including the human skin. Image-based model \cite{levoy1996light,buehler2001unstructured,shum2000review} has been extensively studied in the literature. In particular, the recent image-based rendering solutions \cite{Lawrence21,Virtualcube2022} leveraging advanced depth sensing and deep learning techniques have shown high-quality, real-time rendering of human in the 3D video communication scenario. 

As shown in Fig.~\ref{fig:view_synthesis}, our image-based model is essentially a Lumigraph consisting of a geometry proxy and associated image data. The geometry proxy serves the vital function of determining how to compute the color of each pixel of the rendered image. The geometry proxy we built is a view-dependent one obtained as follows. We start with image and depth data from the RGBD cameras with the background pixels removed. For each view, we computes an initial depth map by fusing the multiview depth maps from the RGBD cameras and then refine it with a multiview-stereo (MVS) neural network~\cite{yao2018mvsnet,luo2019p,Virtualcube2022} using only the RGB images from these cameras. The image data of the Lumigraph are simply the RGB images of the multiple cameras that we place around the display. This Lumigraph is called an Unstructured Lumigraph \cite{buehler2001unstructured} in the literature. 

\begin{figure}[t]
    \centering
    \includegraphics[width=0.92\columnwidth]{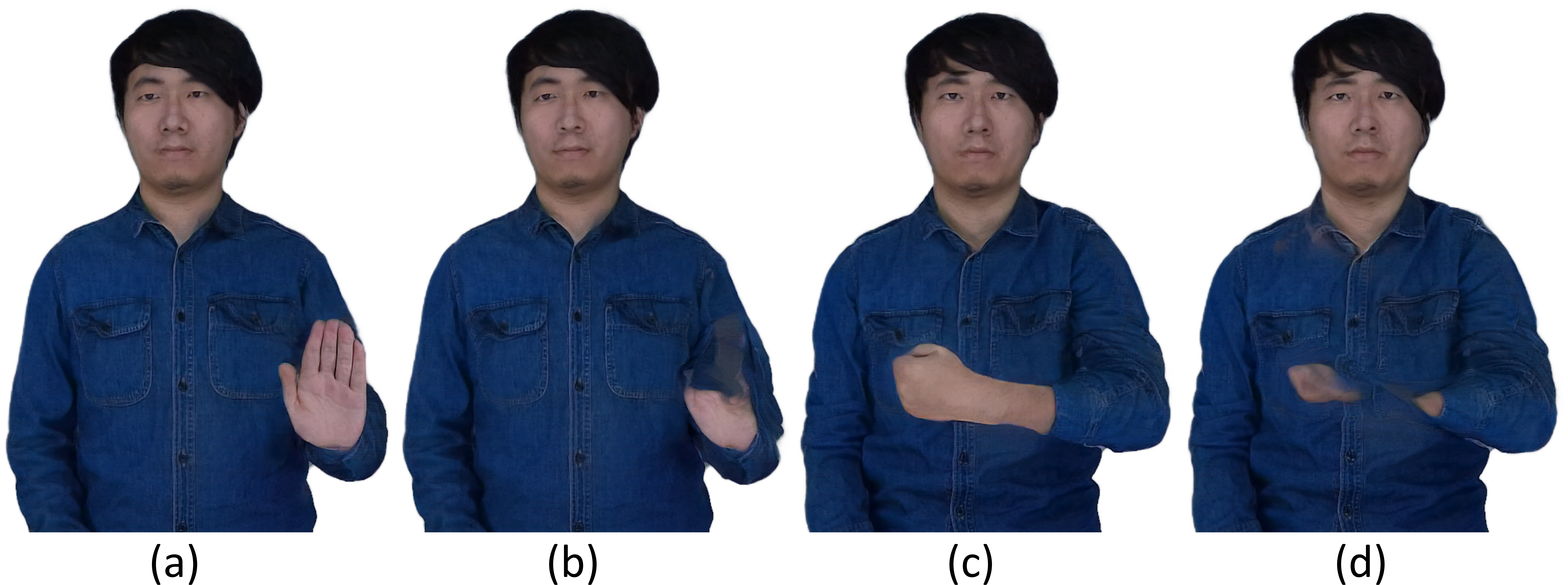}
    \vspace{-0.3cm}
    \caption{
    Comparison of the original rendering method in \cite{Virtualcube2022} and ours. (a) and (c) are rendering results generated by our method and (b) and (d) are images rendered by the original method. It is clear that our method generates better rendering results when the user's hand is in front of his body. 
    }
    \label{fig:depth_init}
\end{figure}

Rendering this Lumigraph for a target view can be done in two steps. First we use the geometry proxy to warp the RGB images of the RGBD cameras to this view. Thus for each pixel in the target view, we have multiple pixel colors from the warped RGB images. The final pixel color is a weighted average of these pixel colors, computed by using the geometry proxy as a guide. This weighted average is usually computed using the conventional Unstructured Lumigraph algorithm~\cite{buehler2001unstructured}. In this work, we compute the weighted average using the neural Lumigraph based rendering method described in \cite{Virtualcube2022}, which uses a neural network called Lumi-Net to compute the weighted average. 

This neural Lumigraph based rendering method generally works well for human body. However, under severe self-occlusion, e.g., when the user waves hand in front of the body or reaches out hand towards the display, the rendering quality of both the hand and the occluded body parts are not satisfactory. We make two algorithmic modifications to improve the quality. The first one is a new edge-preserving depth initialization strategy. We identified that initial depth maps contain blurry edges along depth discontinuity due to the misalignment between the projected multiview depth maps. We replace the averaging operation for depth fusion with a min-filter, which brings obvious hand rendering improvements in practice. Second, we add a new loss function dedicated to the hand region to train the neural networks. Specifically, we detect hand bounding boxes on the target-view images in the training data, and apply a perceptual loss \cite{johnson2016perceptual} between the final network output and the ground truth. Fig.~\ref{fig:depth_init} shows a visual comparison between the original method and ours, which shows the improved hand rendering quality in our system.

Conceptually, one can certainly use other image-based models for our proposed dual representation.

\begin{figure}
    \centering
    \includegraphics[width=0.95\columnwidth]{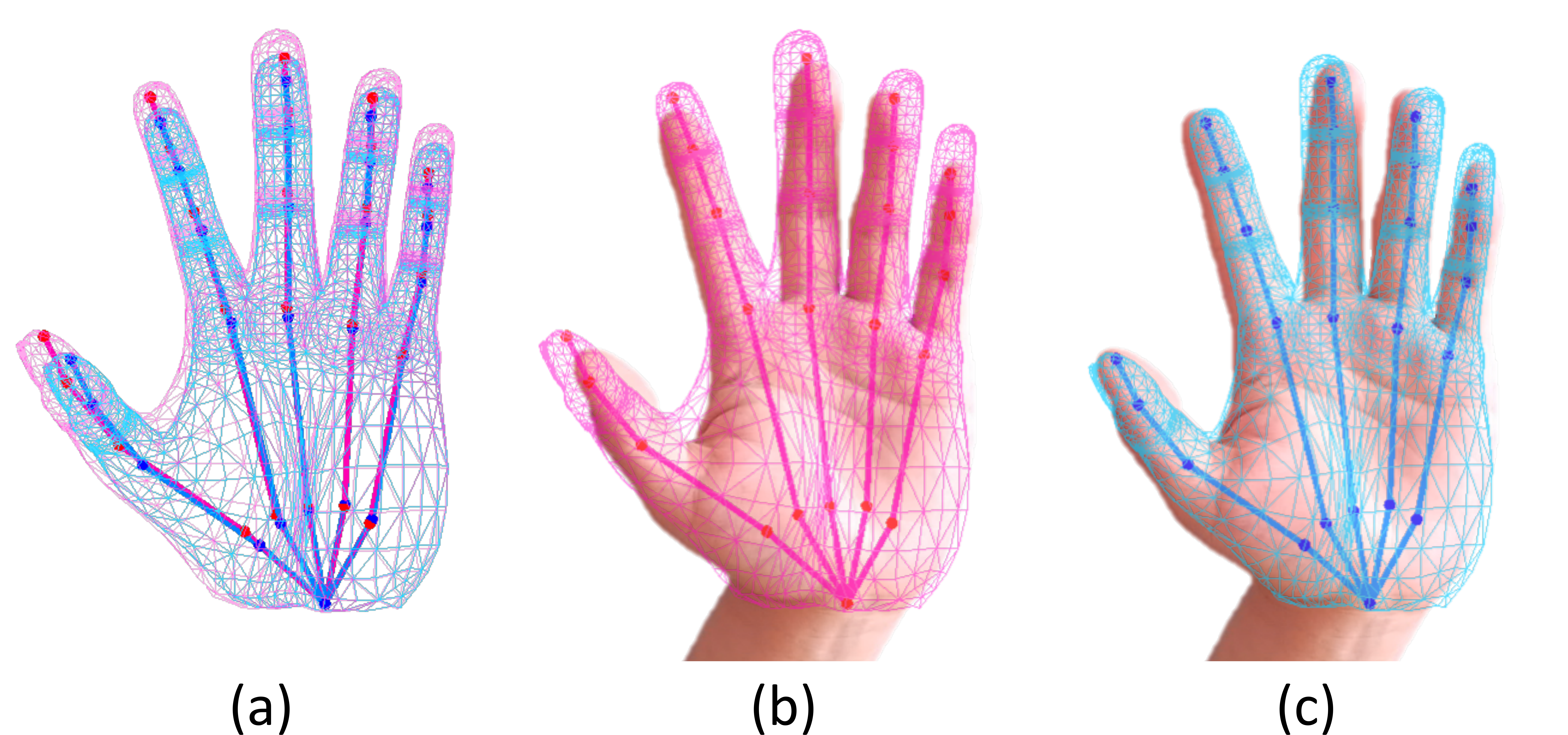}
    \vspace{-0.2cm}
    \caption{
    Shape adaption for model fusion. (a) With the scales computed from the captured skeleton of the real hand, we scale the hand template (in red) with linear blend skinning to obtain the adapted hand model (in blue) (b) Before shape adaption, the shape of the hand template does not match to the user's hand. (c) After shape adaption, the hand mesh fits to the user's hand.
    }
    \label{fig:shape_adaption}
\end{figure}

\subsubsection{3D Geometry-based Model}
As the user's hand moves close to the display, the palm region becomes invisible to all the RGBD cameras. To render the hand palm under such situations, we model the hand with a rigged 3D mesh template.

The 3D hand mesh has a predefined topology and texture map representing the skin surface of a hand at rest pose. It is also associated with a set of 3D skeleton and per-vertex skinning weights with respect to the bones of the skeleton. 

To obtain hand rendering using this mesh model, we track the 3D skeleton of user's real hand by a hand tracking camera, which can successfully reconstruct the position and pose of the real hand as it moves close to the screen and touches the display. The hand mesh and tracked 3D skeleton share the same bone structure so that the former can be easily driven by the latter. Given the local hand motion (bone transformations) reconstructed by the 3D skeleton tracked in each frame, we first rig the hand template mesh using standard linear blending skinning~\cite{lewis2000pose}. Let $V_i$ be a vertex on the rest mesh, its transformed potion after skinning is calculated as 
\begin{equation}
	V_i^{\prime} = \sum_{j} \omega_{ji} \mathbf{T}_j V_i
    \label{eq:lbs}
\end{equation}
where $\mathbf{T}_j$ is the tracked rigid transformation of bone $j$ and $\omega_{ji}$ is the skinning weight of vertex $v_i$ with respect to bone $j$. This way, the skin deforms in a smooth manner. After rigging, we transform the mesh to the real hand's 3D position in the global virtual environment and render it from the target viewpoint of the remote user. 

In our current implementation, we captured the skeleton of the medium-sized hands of a user via LeapMotion and asked an artist to manually modeled the hand mesh and associated skinning weights based on the captured skeleton.

The texture map and hand shape of the mesh model are not fixed, but are adjusted at runtime to fit the user's real hand. This process will be described the next section.

\subsubsection{Model Fusion}\label{sec:model_fusion}

Given two hand representation models, we seamlessly combine them to form our full model and get the final rendering. 

\paragraph{Shape Adaption}
Prior to touching actions, we first adjust the shape of the hand template to fit the shape of the user's real hand. Given the skeleton of the real hand that share the same topology of the skeleton of the hand template at the same pose, we first scale each bone of the template skeleton so that the length of each bone matches the target skeleton. We then update the transformation matrix $B_j$ that transform the local coordinate system of each bone to the global coordinate system due to the changed bone lengths. After that, we scale the hand mesh template by transforming each vertex $V_i$ to its new position $V_i^{\prime}$:
\begin{equation}
	V_i^{\prime} = \sum_{j} \omega_{ji} \mathbf{B}_j^\prime \mathbf{S}_j \mathbf{B}_j^{-1} V_i,
    \label{eq:lbs_scale}
\end{equation}
where $\omega_{ji}$ is the skinning weight of the vertex $V_i$ with respect to the bone $j$. $\mathbf{S}_j$ is a uniform scaling matrix determined by the scaling factor of the bone $j$. Fig.~\ref{fig:shape_adaption} shows the hand mesh before and after shape adaption.

\begin{figure}
    \centering
    \includegraphics[width=0.95\columnwidth]{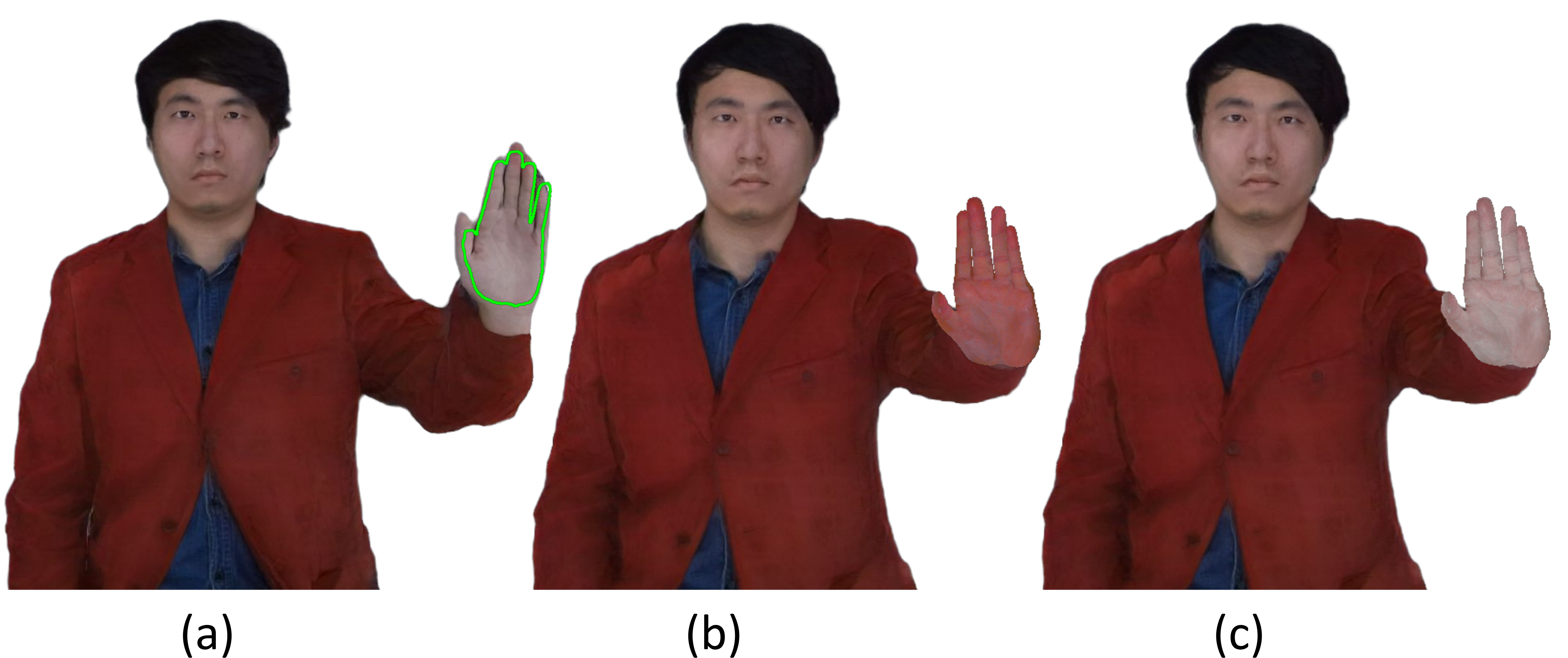}
    \vspace{-0.2cm}
    \caption{
    Appearance adaption for model fusion. (a) The image generated by image-based body and hand rendering. The mask is the projection of 3D geometry-based hand model. (b) The rendering of the geometry-based hand model without appearance adaption. The color of the hand differs a lot from the real hand. (c) With appearance adaption, the rendering of hand is much closer to the real hand.
    }
    \label{fig:3Dhandrendering}
\end{figure}

\paragraph{Appearance Adaption}
Prior to touching actions, we first personalize the mesh model by adjusting its appearance to fit the renderings of the image-based model.
This is an automatic process done online whenever the 3D skeleton of the hand is tracked and the hand palm is rendered by the image-based method while it is well within the field of views of the RGBD cameras (\emph{e.g.}, in the touch starting phase).
To fit the appearance to the rendered hand palm, we adjust the overall color of the hand template texture according to the hand palm image rendered from the image-based model. 
Specifically, let the color and depth images rendered by the image-based model be $\mathbf{I}_i$ and $\mathbf{D}_i$. We can also get the color and depth images of the 3D geometry-based model $\mathbf{I}_g$ and $\mathbf{D}_g$ by rendering the mesh template under the tracked 3D pose using the same camera projection parameters of the image-based model. The region with consistent depth values on $\mathbf{D}_i$ and $\mathbf{D}_g$ (depth difference $<5cm$) is the hand region on the color images, labeled as $\mathbf{I}_i^H$ and $\mathbf{I}_g^H$. This will help to filter out small misalignment pixels. We minimize the color difference between $\mathbf{I}_i^H$ and $\mathbf{I}_g^H$ by calculating a color transformation inspired by \cite{reinhard2001color}, then apply this color transformation to refine the texture $\mathbf{T}$ of the hand mesh model. In this way, the 3D geometry-based model could have similar visual experience with the image-based model.

Specifically, we first convert $\mathbf{I}_i^H$ and $\mathbf{I}_g^H$ to the $L\alpha\beta$ color space. For each channel of $\mathbf{I}_g^H$, we can transform the color $c_g$ of each pixel to match $\mathbf{I}_i^H$ via
\begin{equation}
	c_g^\prime = \frac{\sigma_i}{\sigma_g}(c_g - m_g) + m_i
\end{equation}where $m_i$, $m_g$ and $\sigma_i$, $\sigma_g$ are the mean and standard deviation values of $\mathbf{I}_i^H$ and $\mathbf{I}_g^H$ on the channel. After this color transformation, the resultant image $\mathbf{I}_g^{H \prime}$ has the same mean and standard deviation with $\mathbf{I}_i^H$, making them look similar when transformed back to the RGB color space. As $\mathbf{I}_g^H$ is produced by rendering the 3D hand mesh with texture $\mathbf{T}$ without lighting, the same color transformation can be applied to $\mathbf{T}$ as well, which gives rise to the adjusted texture $\mathbf{T^\prime}$. With this texture, the rendered hand using the 3D hand model will have similar visual appearance with the hand of the image-based model. Fig. ~\ref{fig:3Dhandrendering} illustrates the rendered hand mesh before and after our shape and appearance fitting. Visually inspected, the fitted one well matches the real hand.

\begin{figure}
    \centering
    \includegraphics[width=\columnwidth]{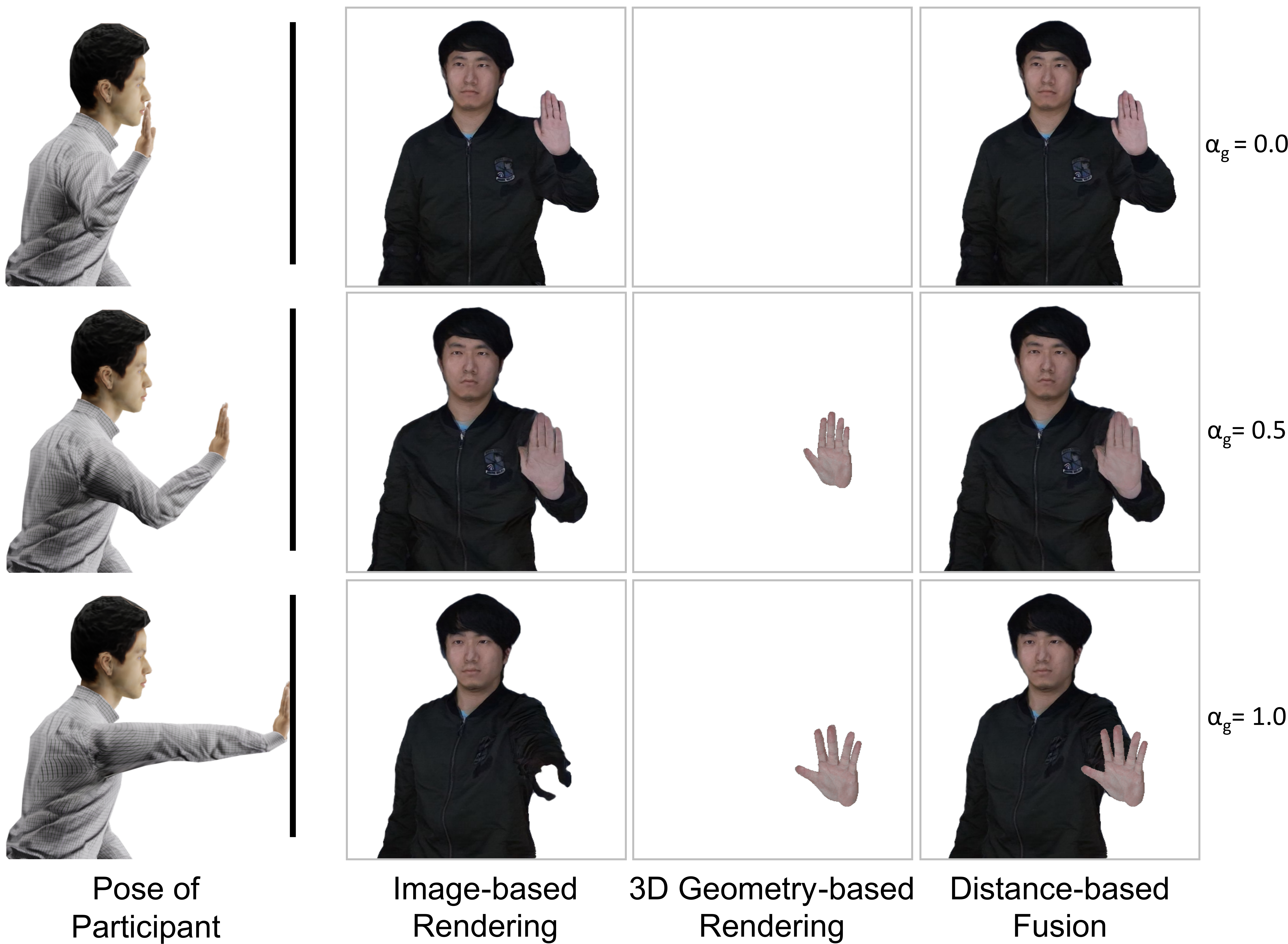}
    \caption{
    The process of distance-based hand rendering fusion. As the hand approaches the screen, the hand skeleton will be detected and rendered and get more weight in the fused image. When the hand is very close to the screen and no longer visible in RGBD cameras, hand is not visible in image-based rendering and mesh-based rendering will be selected.
    }
    \label{fig:touch_fusion}
\end{figure}

\paragraph{Distance-based Rendering Fusion} Given two renderings $\mathbf{I}_i$ and $\mathbf{I}_g$ from the two models, we fuse them to get the final rendering $\mathbf{I}_f$ according to the distance of hand to screen, denoted as $d$. 
Note that $\mathbf{I}_i$ is an RGB$\alpha$ image predicted by the neural network in our image-based model. For $\mathbf{I}_g$ which is the rendering of the 3D hand mesh, we also append it with a distance-based alpha channel to blending it with $\mathbf{I}_i$. Specifically, we calculate the alpha value $\alpha_g$ of a pixel on $\mathbf{I}_g$ as
\begin{equation}
	\mathbf{\alpha}_g = 
	\begin{cases}
		(d_{max} - \hat{d}) / (d_{max} - d_{min})  & \text{for a hand pixel} \\
		0                                          & \text{for a non-hand pixel}\label{eq:alpha_g}
	\end{cases}
\end{equation}
where $\hat{d}=clip(d, [d_{min}, d_{max}])$ is the hand distance clipped to a predefined range $[d_{min}, d_{max}]$. 

To blend the two RGB$\alpha$ images, we set $\mathbf{I}_i$ as the base layer and blend $\mathbf{I}_g$ onto it, as the hand 3D model is in front of the portrait image. The overlay blending mode \cite{wallace1981merging} is used here. Specifically, the alpha and color value ($\mathbf{\alpha}_f$, $\mathbf{c}_f$) of a pixel on the final image $\mathbf{I}_f$ is calculated using the following equations:
\begin{equation}
	\begin{split}
		&\mathbf{\alpha}_f = \mathbf{\alpha}_g + \mathbf{\alpha}_i (1-\mathbf{\alpha}_g), \\
		&\mathbf{c}_f=\frac
		{\mathbf{c}_g \mathbf{\alpha}_g + \mathbf{c}_i \mathbf{\alpha}_i (1 - \mathbf{\alpha}_g)}
		{\mathbf{\alpha}_f}\label{eq:blend}
	\end{split}
\end{equation}where ($\mathbf{\alpha}_g$, $\mathbf{c}_g$) and ($\mathbf{\alpha}_i$, $\mathbf{c}_i$) are alpha and color values of the corresponding pixels on $\mathbf{I}_g$ and $\mathbf{I}_i$, respectively. 

As we can see from Eq.~\eqref{eq:alpha_g} and \eqref{eq:blend}, when $d\geq d_{max}$, we have $\alpha_g=0$, $\alpha_f=\alpha_i$ and $\mathbf{c}_f = \mathbf{c}_i$, \emph{i.e.}, the final rendering simply takes the output of the image-based model. When $d\leq d_{min}$, we have $\alpha_g=\alpha_f=1$ and $\mathbf{c}_f = \mathbf{c}_g$ for the hand region, \emph{i.e.}, the mesh-based rendering is selected. For $d_{min} < d < d_{max}$, which predominately happens in the approaching phase of a touch process, a smooth transition takes place. The fusion process is visualized in Fig.~\ref{fig:touch_fusion}.

In practice, we obtain $d$ for each frame based on the averaged joint positions obtained by the 3D hand skeleton tracker, and we empirically set $d_{max} = 0.4m$ and $d_{min} = 0.2m$ in our system. 
As the final rendering result $\mathbf{I}_f$ contains alpha channel as well, we can compose the hybrid human portrait image with the rendering of a 3D background scene very easily.

\subsection{Touch Emulation}
\label{sec:handdetection}

Beyond visual interaction, our system provides a physical experience of hand touching. We aim to mimic real-world situations where people's hands can receive physical feedback when they touch each other. While the types of touching are various and the corresponding tactile sense are different, we emulate a simple yet commonly-seen scenario, \emph{i.e.}, hand clap, where the participants feel an impact force and vibration when two hands slap each other.

\paragraph{Physical and Virtual Design} To emulate physical touch, we use the screen as the physical media for hand contact. During communication, the two users can reach out their hands to the screens and gently clap each other's rendered hands. Apart from the natural sense of touch when the hand reaches the screen, we provide additional physical feedback when mutual touch is detected.

\begin{figure}
    \centering
    \includegraphics[width=\columnwidth]{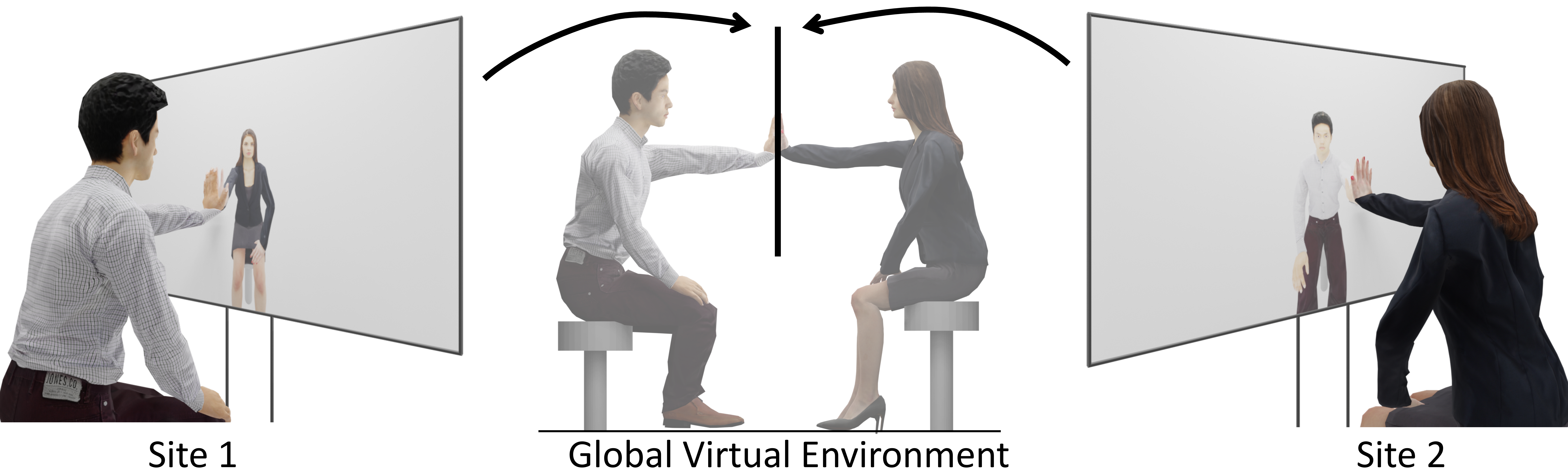}
    \caption{
    The transformation between the physical space and the global virtual space. The position of TV screen from two sites should be aligned in the virtual environment.
    }
    \label{fig:coordinate_transform}
\end{figure}

To realize such a process, we found that maintaining visual-physical consistency is crucial. When the users reach out their hands towards each other, the real and rendered hands on each site should finally coincide and mutual touch should be triggered. This requires a seamless integration of the virtual and physical spaces. 
To achieve this, we transform the physical spaces of two sites into a unified global virtual environment where the surfaces of the two screens overlap, as shown in Fig.~\ref{fig:coordinate_transform}. Under this setup, the content shown on the screen of each site should be a projection of the virtual space behind the screen, as if the users were seeing and performing hand clapping with each other through one transparent glass between them.

It should be noted that since the screen displays are involved in the touch emulation process and their positions matter, our physical and virtual space design described above is different from existing 3D video communication systems such as \cite{Virtualcube2022,Lawrence21} which have no considerations regarding visual-physical consistency. In these systems, the displays are simply visualization portals and they do not necessarily align with each other in the virtual space (the screen positions of \cite{Virtualcube2022} are determined by other factors such as distance between users; the system of \cite{Lawrence21} places the display on each site at the remote user's position).

\paragraph{Mutual Touch Detection and Feedback}
We define mutual touch as the situation where the users' hand palms reach the screen and overlap each other's rendered hand. Note that if the real and rendered hands do not overlap for some reason, we do not call it mutual touch even if both palms are on the screen. 
Our system detects mutual touch by calculating the overlap of two hands. 
Specifically, each site keeps transferring the tracked 3D hand skeleton to the remote site.
For each site, we identify the joints of the local and remote hand skeletons that are within a distance threshold to the screen, then project them onto the screen plane to get two bounding boxes, and finally compute the area of bounding box overlap. If the overlapping area is larger than a threshold, a mutual touch is detected. In our implementation, the joint-to-screen distance threshold is set as 2$cm$ and overlapping area threshold is $50cm^2$.
In practice, this mutual touch detection method can be replaced by other simple alternatives, such as hardware-assisted touch area identification.

Once mutual touch is detected, the system immediately provides additional physical feedback to mimic the feeling of hand clapping. Specifically, a vibration motor tightly attached to the display (Fig.~\ref{fig:system_hardware}) is triggered which incurs screen vibration that can be felt by hand.
Despite of the gap between our simulated vibration and the real sense of impact force and vibration in reality, we found this simple design has already improved the immersive experience and successfully delivers a sense of touch.

%====================================================================================================
%====================================================================================================
\subsection{Implementation Details}

%====================================================================================================

In this section, we present the details of our system including hardware setup and calibration as well as software implementation.

\paragraph{Hardware Setup}
\begin{figure}
	\centering
	\includegraphics[width=\columnwidth]{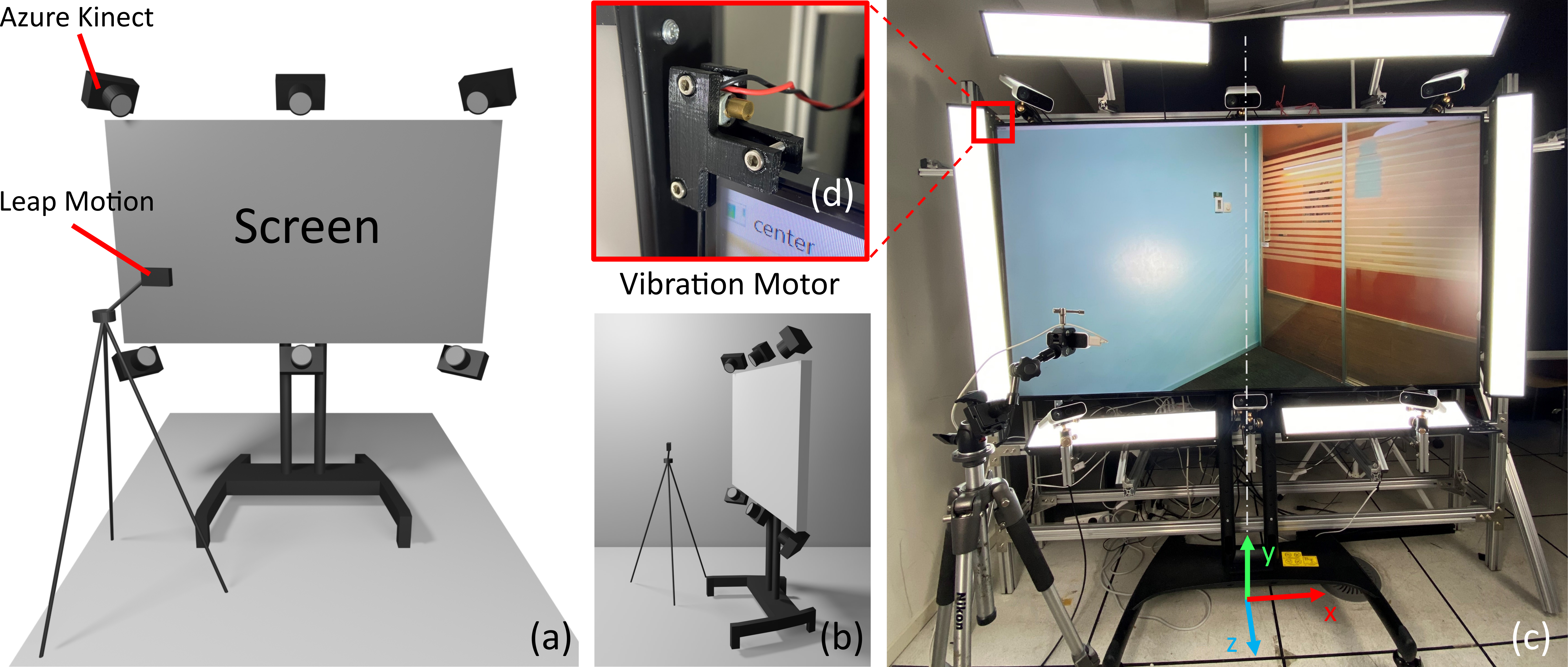}
	\caption{
		The hardware setup of our system. (a) and (b) illustrates the device layout in our design. (c) is an implementation of our design. (d) shows the vibration motor attached to the corner of the screen.  
	}
	\label{fig:system_hardware}
\end{figure}
As shown in Fig.~\ref{fig:system_hardware}, 
our system on each site is equipped with a 65-inch LCD display and six Azure Kinect RGBD cameras. The cameras are installed on a stand behind the display and their positions are adjusted to be near the corners and the center of top and bottom boundaries of the display.
A vibration motor controlled by Arduino is rigidly attached to the top-left corner 
of the screen for offering touch feedback. During remote communication and hand touch, a user sits on a seat that is about $0.7m$ away from the display center. A Leap Motion camera is placed besides the hand for 3D hand tracking. The camera is about $0.5m$ away from the screen and faced to the center of the screen region to which the user's hand can easily touch. In this way, the Leap Motion camera can accurately track the user's hand pose as it approaches and touches the screen. All these devices are connected to a PC and controlled by our system.

%====================================================================================================
\paragraph{Device Calibration}

We calibrate the devices at each site after installation so that all devices and the captured data (\emph{i.e.}, the RGBD frames and 3D hand skeleton of the participant) are all aligned in a unified local coordinate system, which is illustrated in Fig.~\ref{fig:system_hardware} (c).

We calibrate Azure Kinect cameras using the method proposed in \cite{zhang2000flexible}. After that, we manually measure the distance of each Kinect camera to the screen corners and compute a transformation to align the calibrated six Kinect cameras into the local coordinate system from the measured distances. For the Leap Motion camera, we follow the method in \cite{Ahuja2021} for relative pose calibration. Specifically, we render a regular grid of 2D points on the screen with known 3D positions $p^{i}$ in the local coordinate system and then use the index finger of a hand to touch each point in a predefined order and record the 3D positions $q^{i}$ of the finger tip defined in the Leap Motion camera space. The transformation $T$ from the Leap Motion camera space to the local coordinate system can then be computed by $T=\arg\min_{T}{\Sigma_{i}||p^i - T*q^i||}$.

\paragraph{Viewpoint Computation}
To compute the user's viewpoint defined in the local coordinate system, we detect the eye positions in the RGB frames captured by multiple Kinect cameras using the method of \cite{chen2014joint}. The 3D positions can then be computed from their 2D eye positions in the multiple views by triangulation. After that, our method takes the middle point of two eyes as the user's 3D viewpoint. To obtain the target viewpoints for rendering in the local coordinate systems, we transform the detected local user viewpoints of the two sites to the global virtual environment and then transform them to each other's local coordinate system.

%====================================================================================================
\paragraph{Software Workflow}

\begin{figure}
	\centering
	\includegraphics[width=\columnwidth]{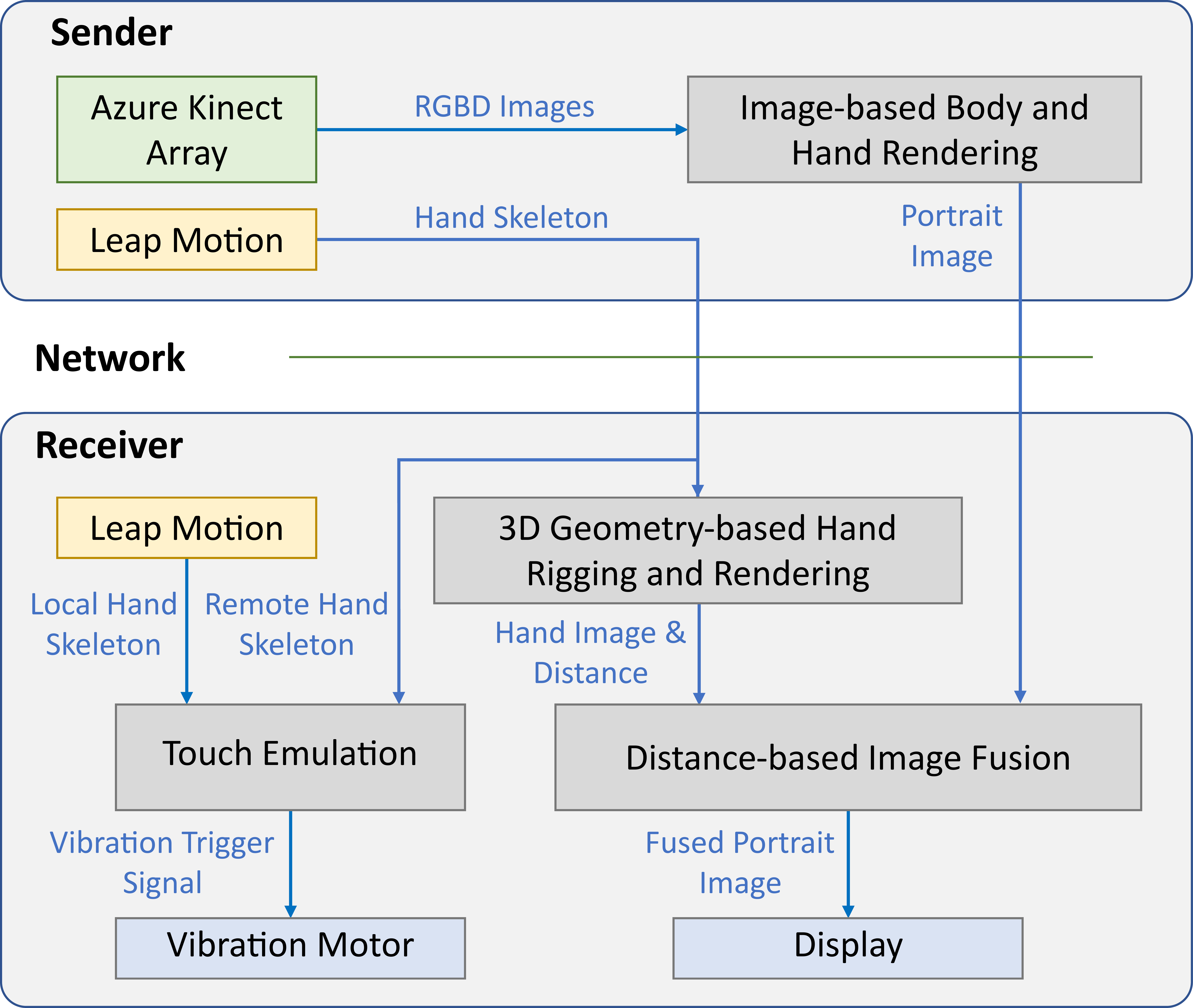}
	\caption{
		The workflow of our RemoteTouch system, which consists of a sender and a receiver at each site. The colorful blocks refer to input and output devices and the grey blocks are software components in our system.
	}
	\label{fig:pipeline}
\end{figure}

The workflow of our system in action is shown in Fig.~\ref{fig:pipeline}. In our system, each site has a sender and a receiver. During online communication, the sender first computes the local user's viewpoint and transmit it to the remote site. It then captures the RGBD frames and 3D hand skeleton of the user and then renders the portrait (\emph{i.e.}, body and hand) image at the received target viewpoint based on image-based representation described in Section~\ref{sec:imagehand}. After that, the sender transmits the rendered portrait images and 3D hand skeleton to the receiver at the remote site. The receiver animates the geometry-based 3D hand model with the received 3D hand skeleton and then renders its image. Finally, the receiver performs the distance-based fusion to blend the hand images rendered from two representations (Section~\ref{sec:handrendering}) and composes the rendering results with the image of the 3D background scene to generate the final image. Meanwhile, the receiver detects mutual hand touch of two users with the algorithm described in Section~\ref{sec:handdetection} and triggers the vibration motor once mutual hand touch happens.
Our system executes all software modules (\emph{i.e.} image-based rendering, hand tracking, geometry-based hand rendering, and touch detection) in different threads in parallel on each site.

\section{Experiments}

We implemented two instances of our system at two sites, each with a PC that has an Intel Core i9-10980XE CPU, 64GB memory and two Nvidia GeForce RTX 3090 GPUs. The two PCs are connected to the LAN network with 1Gbps bandwidth. 
During communication, our method takes RGBD frames of four Azure Kinects that are on the opposite side of touching hand for rendering (\emph{e.g.}, two central and two right ones for the left hand). This camera configuration can minimize the occlusion region caused by the raised hand in the captured frames and thus generates better rendering results than other choices. For viewpoint computation, we assume the user's head does not move a large distance during the communication and thus compute the user's viewpoint only once at the beginning of the communication.

%====================================================================================================
%====================================================================================================
\paragraph{System Performance.}
For image-based body and hand representation, the Azure Kinect cameras capture synchronized RGBD frames at 30fps, with RGB and depth at $2560\times1440$ and $512\times512$ resolutions respectively. The image-based modeling and rendering algorithm takes nearly 150ms to process one frame (from raw data captured by Azure Kinect to synthesized portrait image sending to network). Since each step in the algorithm can be executed in parallel, the system renders the portrait and hand image at 30FPS. The end-to-end delay of portrait image rendering and transmission is around 400ms.

For geometry-based hand representation, the Leap Motion tracks hand skeleton around 50Hz and the 3D hand rigging and rendering can be performed on GPU in real time. The total delay of the geometry-based hand rendering and network transmission is around 250ms. We synchronize the rendering and display of these two representations with a timestamp on each frame and fuses the images of two representations with the closest timestamp in the receiver at 30FPS. 

For touch emulation, the touch detection is performed at 50Hz. The delay from the vibration triggering to physical motor vibration is around 60ms, which is hardly noticed by users. 

\begin{figure*}[t]
    \centering
    \includegraphics[width=\textwidth]{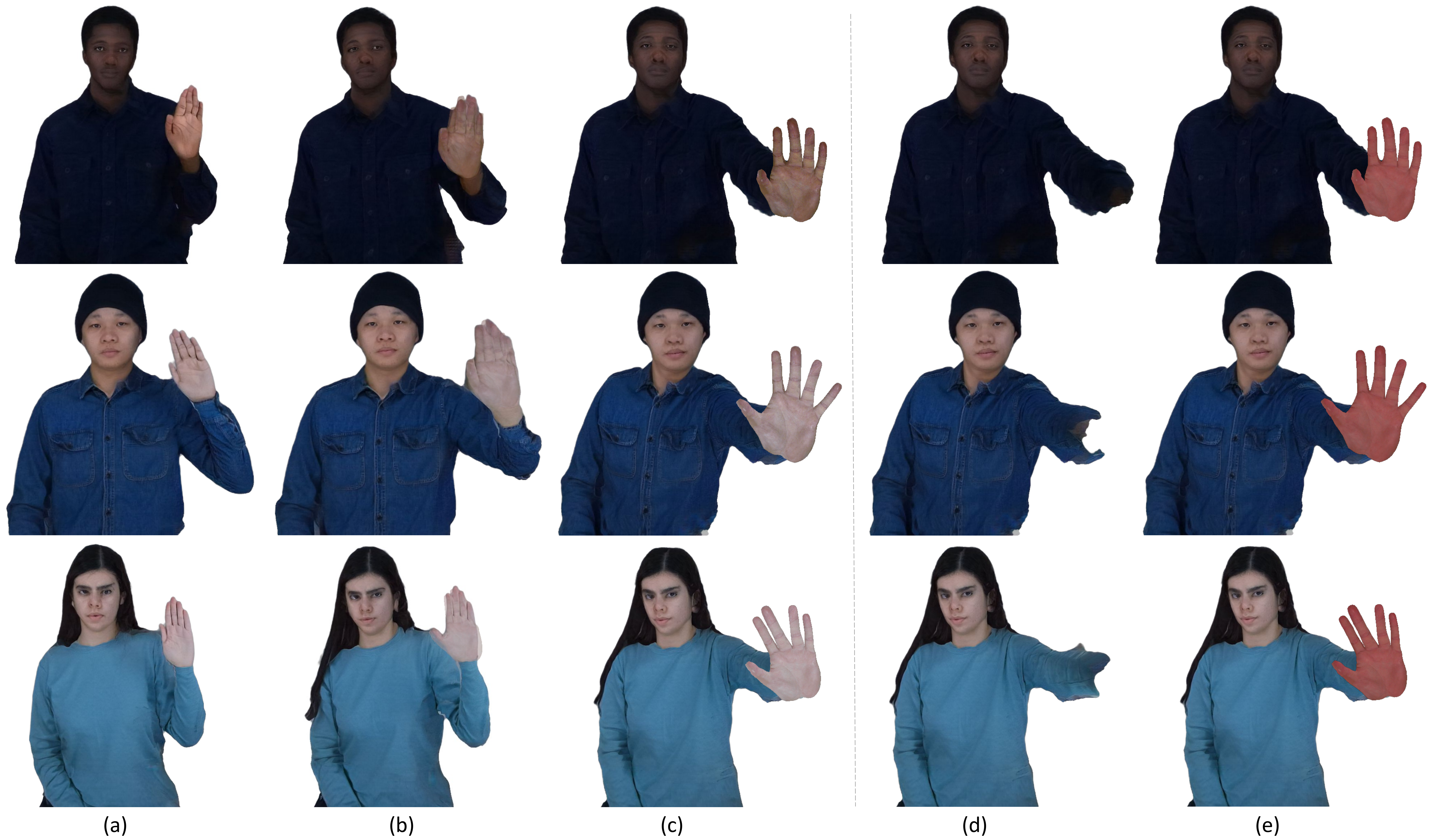}
    \caption{
    The rendering of the subjects with different skin tones, where the hand are rendered by image-based model (a), fused dual representation (b), and geometry-based model (c) where the hand touched the screen. Our dual representation and fusion scheme generate convincing results for different poses. When hand touch the screen, the hand is not correctly rendered by the image-based model (d) as the hand is invisible in source view cameras. If appearance adaption is not applied to geometry-based model (e), the original color of the hand differs a lot from the hand.
    }
    \label{fig:results}
\end{figure*}

%====================================================================================================
%====================================================================================================
\paragraph{Evaluation of Dual Hand Representation.}
Fig.~\ref{fig:results} (a-c) illustrates the rendering frames of the participants sampled in different hand touch phases. Note that our dual representation and model fusion algorithm generates convincing rendering results as the hand is at different distance to the screen. When the hand touch the screen, the palm region is missing in the image-based rendering result, as the hand is invisible to source view cameras, as shown in Fig.~\ref{fig:results} (d). In this case, our dual representation of hand successfully render the final portrait image by fusing the 3D geometry-based model into the image-based rendering, as shown in Fig.~\ref{fig:results} (c). Fig.~\ref{fig:results} (e) show the hand without appearance adaption to the geometry-based model, which looks unnatural compared with other parts of the body. Our appearance fitting scheme is robust to different skin tones.

%====================================================================================================
%====================================================================================================

\paragraph{User Experience.}

We invited 26 participants to try and evaluate our touch-enhanced 3D video communication system. All of them have experience on commercial video conferencing software on PCs or cellphones but are the first-time user of our system. During the experiment, our developer sits in one site and the user sits in the other site. After a short introduction, the two users can chat with each other and try the hand clapping via our system. After the experiment, we ask each user to provide comments on our system and rank the impact of touch experience to remote video communication. Among all 26 users, 24 of 26 ($92\%$) participants agreed or strongly agreed that the remote touch enhances the experience and decreases mental distance between the remote users.

To further evaluate the contribution of our dual hand representation to the user experience, we let all 26 participants to try the hand clapping rendered with image-based representation only and with our dual-hand representation. Our investigation demonstrated that all users agreed that our dual hand representation enhances the touch experience compared with image-based representation alone, even if they noticed the artifacts of hand misalignment when their hand approached to the screen. Several users with computer graphics background mentioned that they expected better portrait image quality, which may significantly improve their experience. They also expected better hand tracking accuracy and less system latency. Despite these shortcomings, all users still liked the feeling that they can visualize the hand and arm in the full hand clap process. They never have such experience in other remote communication applications.

Finally, we asked the user to evaluate the impact of vibration for touch experience by testing the hand clapping with vibration and without vibration feedback. 25 of 26 ($96\%$) participants commented that the vibration makes the hand clapping more attractive. One user mentioned that the vibration made her believe that the touch did happened. 7 users mentioned that they are excited about the idea haptic feedback when touching and expect more kinds of physical feedback, such as the sound of clapping hands. 5 users expect more kinds hand touching gestures to be supported, such as a fist bump.

\section{Conclusion and Future Work}\label{sec:discussions}

We present a method for emulating hand touch (“the high five”) between two remote participants. The key of our method is a dual representation of hand that combines the advantages of the image-based and 3D geometry-based representation and a distance-based fusion scheme for rendering. We also design a touch emulation scheme that map the two remote site into one shared environment. Our solution offers a visual-physical consistent touch experience for immersive 3D video communication. We validate our dual representation for hand, and our experiments demonstrate that our method reduces the mental distance between the remote users and enhances the user experience in immersive 3D video communication. 

Currently, our method only supports hand touch between two remote users. 
Extending our method to hand touch between three or more remote users would be an interesting topic worth further exploration.

The misalignment between geometry-based hand and real hand in our prototype is mainly due to the inaccurate 3D hand tracking provided by the LeapMotion. A more accurate real-time hand tracking method will resolve this issue. For model fusion, our method adapts the hand shape according to the bone length and fails to compensate the detailed shape difference between the real hand and hand mesh template. Also we use the same texture map to approximate the detailed appearance of different hands. It is interesting to explore how to capture and reconstruct the detailed geometry and texture of the real hands on the fly.

We also would like to further improve the hardware and software design of our method to facilitate better touch experience in 3D video communication, which includes simplifying the device setup for data capturing, integrating other force feedback devices, and reducing the system delay for rendering and network transmission. Besides these improvements, a possible application of our method is the collaborative whiteboard system. Another possible research direction is to enable more social touch behaviors in immersive 3D video communication.

\subsection*{Acknowledgment}
We would like to thank Yuxiao Guo, Guojun Chen for neural network conversion with NNFusion, Yangyu Huang for optimizing the training and run-time code of neural lumigraph algorithm, Haowen Xu for his support in camera calibration, all colleagues for their help in data capturing and useful discussions, and our users for their valuable feedback of our RemoteTouch system. We also thank the anonymous reviewers for their helpful suggestions for our paper.

\bibliographystyle{abbrv-doi}

\bibliography{vc}

\begin{thebibliography}{10}

\bibitem{leapmotion}
Ultraleap, how hand tracking works.
\newblock
  \url{https://www.ultraleap.com/company/news/blog/how-hand-tracking-works/}.

\bibitem{Ahuja2021}
K.~Ahuja, P.~Streli, and C.~Holz.
\newblock Touchpose: Hand pose prediction, depth estimation, and touch
  classification from capacitive images.
\newblock In {\em UIST}, p. 997–1009. ACM, 2021.

\bibitem{alsamarei_2021}
A.~A.~A. Alsamarei.
\newblock {\em Remote social touch: a framework to communicate physical
  interaction across long distances}.
\newblock PhD thesis, Middle East Technical University, 2021.

\bibitem{baker2002coliseum}
H.~Baker, D.~Tanguay, I.~Sobel, D.~Gelb, M.~E. Goss, W.~B. Culbertson, and
  T.~Malzbender.
\newblock The coliseum immersive teleconferencing system.
\newblock In {\em International Workshop on Immersive Telepresence}, vol.~6,
  2002.

\bibitem{Benko2005}
H.~Benko, E.~Ishak, and S.~Feiner.
\newblock Cross-dimensional gestural interaction techniques for hybrid
  immersive environments.
\newblock In {\em IEEE Proc. VR}, pp. 209--216, 2005.

\bibitem{HapticAR2021}
C.~Bermejo and P.~Hui.
\newblock A survey on haptic technologies for mobile augmented reality.
\newblock {\em ACM Comput. Surv.}, 54(9), oct 2021.

\bibitem{bevan2015shaking}
C.~Bevan and D.~Stanton~Fraser.
\newblock Shaking hands and cooperation in tele-present human-robot
  negotiation.
\newblock In {\em Proceedings of the tenth annual ACM/IEEE international
  conference on human-robot interaction}, pp. 247--254, 2015.

\bibitem{HapticVR2021}
E.~Bouzbib, G.~Bailly, S.~Haliyo, and P.~Frey.
\newblock “can i touch this?”: Survey of virtual reality interactions via
  haptic solutions: Revue de litt\'{e}rature des interactions en
  r\'{e}alit\'{e} virtuelle par le biais de solutions haptiques.
\newblock In {\em 32e Conf\'{e}rence Francophone Sur l'Interaction
  Homme-Machine}. ACM, 2021.

\bibitem{buehler2001unstructured}
C.~Buehler, M.~Bosse, L.~McMillan, S.~Gortler, and M.~Cohen.
\newblock Unstructured lumigraph rendering.
\newblock In {\em SIGGRAPH}, pp. 425--432, 2001.

\bibitem{Burdea1992}
G.~Burdea, J.~Zhuang, E.~Roskos, D.~Silver, and N.~Langrana.
\newblock A portable dextrous master with force feedback.
\newblock {\em Presence: Teleoper. Virtual Environ.}, 1(1):18–28, 1992.

\bibitem{Buxton1997}
W.~Buxton, A.~Sellen, and M.~Sheasby.
\newblock Interfaces for multiparty videoconferences.
\newblock 1997.

\bibitem{Carter2013}
T.~Carter, S.~A. Seah, B.~Long, B.~Drinkwater, and S.~Subramanian.
\newblock Ultrahaptics: Multi-point mid-air haptic feedback for touch surfaces.
\newblock In {\em UIST}, p. 505–514. ACM, 2013.

\bibitem{chen2014joint}
D.~Chen, S.~Ren, Y.~Wei, X.~Cao, and J.~Sun.
\newblock Joint cascade face detection and alignment.
\newblock In {\em ECCV}, pp. 109--122. Springer, 2014.

\bibitem{Cheng2017}
L.-P. Cheng, E.~Ofek, C.~Holz, H.~Benko, and A.~D. Wilson.
\newblock Sparse haptic proxy: Touch feedback in virtual environments using a
  general passive prop.
\newblock In {\em CHI}, p. 3718–3728. ACM, 2017.

\bibitem{Coldefy2007}
F.~Coldefy and S.~Louis-dit Picard.
\newblock Digitable: an interactive multiuser table for collocated and remote
  collaboration enabling remote gesture visualization.
\newblock In {\em CVPR}, pp. 1--8, 2007.

\bibitem{collet2015high}
A.~Collet, M.~Chuang, P.~Sweeney, D.~Gillett, D.~Evseev, D.~Calabrese,
  H.~Hoppe, A.~Kirk, and S.~Sullivan.
\newblock High-quality streamable free-viewpoint video.
\newblock {\em ACM TOG}, 34(4):1--13, 2015.

\bibitem{dou2016fusion4d}
M.~Dou, S.~Khamis, Y.~Degtyarev, P.~Davidson, S.~R. Fanello, A.~Kowdle, S.~O.
  Escolano, C.~Rhemann, D.~Kim, J.~Taylor, et~al.
\newblock Fusion4d: Real-time performance capture of challenging scenes.
\newblock {\em ACM TOG}, 35(4):1--13, 2016.

\bibitem{gallace2022social}
A.~Gallace and M.~Girondini.
\newblock Social touch in virtual reality.
\newblock {\em Current Opinion in Behavioral Sciences}, 43:249--254, 2022.

\bibitem{GALLACE2010246}
A.~Gallace and C.~Spence.
\newblock The science of interpersonal touch: An overview.
\newblock {\em Neuroscience \& Biobehavioral Reviews}, 34(2):246--259, 2010.
\newblock Touch, Temperature, Pain/Itch and Pleasure.

\bibitem{Genest2013}
A.~M. Genest, C.~Gutwin, A.~Tang, M.~Kalyn, and Z.~Ivkovic.
\newblock Kinectarms: A toolkit for capturing and displaying arm embodiments in
  distributed tabletop groupware.
\newblock In {\em Proceedings of the 2013 Conference on Computer Supported
  Cooperative Work}, p. 157–166. ACM, 2013.

\bibitem{gibbs1999teleport}
S.~J. Gibbs, C.~Arapis, and C.~J. Breiteneder.
\newblock Teleport--towards immersive copresence.
\newblock {\em Multimedia Systems}, 7(3):214--221, 1999.

\bibitem{gotsch2018telehuman2}
D.~Gotsch, X.~Zhang, T.~Merritt, and R.~Vertegaal.
\newblock Telehuman2: A cylindrical light field teleconferencing system for
  life-size 3d human telepresence.
\newblock In {\em CHI}, vol.~18, p. 552, 2018.

\bibitem{Guo2017}
K.~Guo, F.~Xu, T.~Yu, X.~Liu, Q.~Dai, and Y.~Liu.
\newblock Real-time geometry, albedo, and motion reconstruction using a single
  rgb-d camera.
\newblock {\em ACM TOG}, 36(4), 2017.

\bibitem{haans2006mediated}
A.~Haans and W.~IJsselsteijn.
\newblock Mediated social touch: a review of current research and future
  directions.
\newblock {\em Virtual Reality}, 9(2):149--159, 2006.

\bibitem{Keita2015}
K.~Higuchi, Y.~Chen, P.~A. Chou, Z.~Zhang, and Z.~Liu.
\newblock Immerseboard: Immersive telepresence experience using a digital
  whiteboard.
\newblock In {\em CHI}, p. 2383–2392. ACM, 2015.

\bibitem{huang2021survey}
L.~Huang, B.~Zhang, Z.~Guo, Y.~Xiao, Z.~Cao, and J.~Yuan.
\newblock Survey on depth and rgb image-based 3d hand shape and pose
  estimation.
\newblock {\em Virtual Reality \& Intelligent Hardware}, 3:207--234, 2021.

\bibitem{Ishii1992}
H.~Ishii and M.~Kobayashi.
\newblock Clearboard: A seamless medium for shared drawing and conversation
  with eye contact.
\newblock In {\em CHI}, p. 525–532. ACM, 1992.

\bibitem{Ishii1991}
H.~Ishii and N.~Miyake.
\newblock Toward an open shared workspace: Computer and video fusion approach
  of teamworkstation.
\newblock {\em Commun. ACM}, 34(12):37–50, dec 1991.

\bibitem{Iwai2018}
D.~Iwai, R.~Matsukage, S.~Aoyama, T.~Kikukawa, and K.~Sato.
\newblock Geometrically consistent projection-based tabletop sharing for remote
  collaboration.
\newblock {\em IEEE Access}, 6:6293--6302, 2018.

\bibitem{Izadi2007}
S.~Izadi, A.~Agarwal, A.~Criminisi, J.~Winn, A.~Blake, and A.~Fitzgibbon.
\newblock C-slate: A multi-touch and object recognition system for remote
  collaboration using horizontal surfaces.
\newblock In {\em Second Annual IEEE International Workshop on Horizontal
  Interactive Human-Computer Systems}, pp. 3--10, 2007.

\bibitem{johnson2016perceptual}
J.~Johnson, A.~Alahi, and L.~Fei-Fei.
\newblock Perceptual losses for real-time style transfer and super-resolution.
\newblock In {\em ECCV}, pp. 694--711, 2016.

\bibitem{Jones2009}
A.~Jones, M.~Lang, G.~Fyffe, X.~Yu, J.~Busch, I.~McDowall, M.~Bolas, and
  P.~Debevec.
\newblock Achieving eye contact in a one-to-many 3d video teleconferencing
  system.
\newblock {\em ACM TOG}, 28(3), 2009.

\bibitem{Sasa2012}
S.~Junuzovic, K.~Inkpen, T.~Blank, and A.~Gupta.
\newblock Illumishare: Sharing any surface.
\newblock In {\em CHI}, p. 1919–1928. ACM, 2012.

\bibitem{kauff2002immersive}
P.~Kauff and O.~Schreer.
\newblock An immersive 3d video-conferencing system using shared virtual team
  user environments.
\newblock In {\em International Conference on Collaborative Virtual
  Environments}, pp. 105--112, 2002.

\bibitem{khaleghi2022multi}
L.~Khaleghi, A.~Sepas-Moghaddam, J.~Marshall, and A.~Etemad.
\newblock Multi-view video-based 3d hand pose estimation.
\newblock {\em IEEE Transactions on Artificial Intelligence}, 2022.

\bibitem{Kim2014RetroDepth3S}
D.~Kim, S.~Izadi, J.~Dostal, C.~Rhemann, C.~Keskin, C.~Zach, J.~Shotton, T.~A.
  Large, S.~Bathiche, M.~Nie{\ss}ner, A.~Butler, S.~Fanello, and V.~Pradeep.
\newblock Retrodepth: 3d silhouette sensing for high-precision input on and
  above physical surfaces.
\newblock {\em CHI}, 2014.

\bibitem{kuechler2006}
M.~Kuechler and A.~Kunz.
\newblock Holoport-a device for simultaneous video and data conferencing
  featuring gaze awareness.
\newblock In {\em IEEE VR}, pp. 81--88, 2006.

\bibitem{kuster2012}
C.~Kuster, N.~Ranieri, H.~Zimmer, J.-C. Bazin, C.~Sun, T.~Popa, M.~Gross,
  et~al.
\newblock Towards next generation 3d teleconferencing systems.
\newblock In {\em 3DTV-Conference: The True Vision-Capture, Transmission and
  Display of 3D Video}, pp. 1--4, 2012.

\bibitem{Lawrence21}
J.~Lawrence, D.~B. Goldman, S.~Achar, G.~M. Blascovich, J.~G. Desloge,
  T.~Fortes, E.~M. Gomez, S.~Häberling, H.~Hoppe, A.~Huibers, C.~Knaus,
  B.~Kuschak, R.~Martin-Brualla, H.~Nover, A.~I. Russell, S.~M. Seitz, and
  K.~Tong.
\newblock Project starline: A high-fidelity telepresence system.
\newblock {\em SIGGRAPH Asia}, 40(6), 2021.

\bibitem{Le2017}
K.-D. Le, K.~Zhu, and M.~Fjeld.
\newblock Mirrortablet: Exploring a low-cost mobile system for capturing
  unmediated hand gestures in remote collaboration.
\newblock In {\em Proceedings of the 16th International Conference on Mobile
  and Ubiquitous Multimedia}, p. 79–89. ACM, 2017.

\bibitem{Leithinger2014}
D.~Leithinger, S.~Follmer, A.~Olwal, and H.~Ishii.
\newblock Physical telepresence: Shape capture and display for embodied,
  computer-mediated remote collaboration.
\newblock In {\em UIST}, p. 461–470. ACM, 2014.

\bibitem{levoy1996light}
M.~Levoy and P.~Hanrahan.
\newblock Light field rendering.
\newblock In {\em SIGGRAPH}, pp. 31--42, 1996.

\bibitem{lewis2000pose}
J.~P. Lewis, M.~Cordner, and N.~Fong.
\newblock Pose space deformation: a unified approach to shape interpolation and
  skeleton-driven deformation.
\newblock In {\em SIGGRAPH}, pp. 165--172, 2000.

\bibitem{lim2020camera}
G.~M. Lim, P.~Jatesiktat, C.~W.~K. Kuah, and W.~T. Ang.
\newblock Camera-based hand tracking using a mirror-based multi-view setup.
\newblock In {\em 2020 42nd Annual International Conference of the IEEE
  Engineering in Medicine \& Biology Society (EMBC)}, pp. 5789--5793. IEEE,
  2020.

\bibitem{Lindeman2004}
R.~W. Lindeman, R.~Page, Y.~Yanagida, and J.~L. Sibert.
\newblock Towards full-body haptic feedback: The design and deployment of a
  spatialized vibrotactile feedback system.
\newblock In {\em Proceedings of the ACM Symposium on Virtual Reality Software
  and Technology}, p. 146–149. ACM, 2004.

\bibitem{luo2019p}
K.~Luo, T.~Guan, L.~Ju, H.~Huang, and Y.~Luo.
\newblock P-mvsnet: Learning patch-wise matching confidence aggregation for
  multi-view stereo.
\newblock In {\em CVPR}, pp. 10452--10461, 2019.

\bibitem{maimone2011encumbrance}
A.~Maimone and H.~Fuchs.
\newblock Encumbrance-free telepresence system with real-time 3d capture and
  display using commodity depth cameras.
\newblock In {\em ISMAR}, pp. 137--146, 2011.

\bibitem{Massie1994ThePH}
T.~Massie.
\newblock The phantom haptic interface: A device for probing virtual objects.
\newblock 1994.

\bibitem{mehrabian2017nonverbal}
A.~Mehrabian.
\newblock {\em Nonverbal communication}.
\newblock Routledge, 2017.

\bibitem{nakanishi2014remote}
H.~Nakani~shi, K.~Tanaka, and Y.~Wada.
\newblock Remote handshaking: touch enhances video-mediated social
  telepresence.
\newblock In {\em CHI}, pp. 2143--2152, 2014.

\bibitem{nguyen2005multiview}
D.~Nguyen and J.~Canny.
\newblock Multiview: spatially faithful group video conferencing.
\newblock In {\em CHI}, pp. 799--808, 2005.

\bibitem{oh2016hand}
J.~Oh, Y.~Lee, Y.~Kim, T.~Jin, S.~Lee, and S.-H. Lee.
\newblock Hand contact between remote users through virtual avatars.
\newblock In {\em Proceedings of the 29th International Conference on Computer
  Animation and Social Agents}, pp. 97--100, 2016.

\bibitem{Onishi2017SpatialCA}
Y.~Onishi, K.~Tanaka, and H.~Nakanishi.
\newblock Spatial continuity and robot-embodied pointing behavior in
  videoconferencing.
\newblock In {\em CRIWG}, 2017.

\bibitem{Orts2016}
S.~Orts-Escolano, C.~Rhemann, S.~Fanello, W.~Chang, A.~Kowdle, Y.~Degtyarev,
  D.~Kim, P.~L. Davidson, S.~Khamis, M.~Dou, V.~Tankovich, C.~Loop, Q.~Cai,
  P.~A. Chou, S.~Mennicken, J.~Valentin, V.~Pradeep, S.~Wang, S.~B. Kang,
  P.~Kohli, Y.~Lutchyn, C.~Keskin, and S.~Izadi.
\newblock Holoportation: Virtual 3d teleportation in real-time.
\newblock In {\em UIST}, p. 741–754. ACM, 2016.

\bibitem{Pezent2019}
E.~Pezent, A.~Israr, M.~Samad, S.~Robinson, P.~Agarwal, H.~Benko, and
  N.~Colonnese.
\newblock Tasbi: Multisensory squeeze and vibrotactile wrist haptics for
  augmented and virtual reality.
\newblock In {\em 2019 IEEE World Haptics Conference}, pp. 1--6, 2019.

\bibitem{Pluss2016}
C.~Pl\"{u}ss, N.~Ranieri, J.-C. Bazin, T.~Martin, P.-Y. Laffont, T.~Popa, and
  M.~Gross.
\newblock An immersive bidirectional system for life-size 3d communication.
\newblock In {\em International Conference on Computer Animation and Social
  Agents}, p. 89–96. ACM, 2016.

\bibitem{Price2020}
S.~Price, C.~Jewitt, and N.~Yiannoutsou.
\newblock Conceptualising touch in vr.
\newblock {\em Virtual Real.}, 25(3):863–877, sep 2021.

\bibitem{reinhard2001color}
E.~Reinhard, M.~Adhikhmin, B.~Gooch, and P.~Shirley.
\newblock Color transfer between images.
\newblock {\em IEEE Computer graphics and applications}, 21(5):34--41, 2001.

\bibitem{sadagic2001tele}
A.~Sadagic, H.~Towles, L.~Holden, K.~Daniilidis, and B.~Zeleznik.
\newblock Tele-immersion portal: Towards an ultimate synthesis of computer
  graphics and computer vision systems.
\newblock In {\em Annual International Workshop on Presence}, pp. 21--23, 2001.

\bibitem{RemoteCollabSurvery2021}
A.~Sch{\"{a}}fer, G.~Reis, and D.~Stricker.
\newblock A survey on synchronous augmented, virtual and mixed reality remote
  collaboration systems.
\newblock {\em CoRR}, abs/2102.05998, 2021.

\bibitem{Schmitz2021}
M.~Schmitz, F.~M\"{u}ller, M.~M\"{u}hlh\"{a}user, J.~Riemann, and H.~V.~V. Le.
\newblock Itsy-bits: Fabrication and recognition of 3d-printed tangibles with
  small footprints on capacitive touchscreens.
\newblock In {\em CHI}. ACM, 2021.

\bibitem{seinfeld2022evoking}
S.~Seinfeld, I.~Schmidt, and J.~M{\"u}ller.
\newblock Evoking realistic affective touch experiences in virtual reality.
\newblock {\em arXiv preprint arXiv:2202.13389}, 2022.

\bibitem{shum2000review}
H.~Shum and S.~B. Kang.
\newblock Review of image-based rendering techniques.
\newblock In {\em Visual Communications and Image Processing}, vol. 4067, pp.
  2--13, 2000.

\bibitem{sin2013human}
M.~T.~A. Sin, S.~L. Koole, et~al.
\newblock That human touch that means so much: exploring the tactile dimension
  of social life.
\newblock {\em Mind Magazine}, 2:17, 2013.

\bibitem{10.1145/3414685.3417768}
B.~Smith, C.~Wu, H.~Wen, P.~Peluse, Y.~Sheikh, J.~K. Hodgins, and T.~Shiratori.
\newblock Constraining dense hand surface tracking with elasticity.
\newblock {\em TOG}, 39(6), 2020.

\bibitem{Sodhi2013}
R.~Sodhi, I.~Poupyrev, M.~Glisson, and A.~Israr.
\newblock Aireal: Interactive tactile experiences in free air.
\newblock {\em ACM TOG}, 32(4), jul 2013.

\bibitem{Streli2021}
P.~Streli and C.~Holz.
\newblock Capcontact: Super-resolution contact areas from capacitive
  touchscreens.
\newblock In {\em CHI}. ACM, 2021.

\bibitem{Sykownik2020}
P.~Sykownik and M.~Masuch.
\newblock The experience of social touch in multi-user virtual reality.
\newblock In {\em 26th ACM Symposium on Virtual Reality Software and
  Technology}. ACM, 2020.

\bibitem{Tan2009}
K.-H. Tan, I.~Robinson, R.~Samadani, B.~Lee, D.~Gelb, A.~Vorbau, B.~Culbertson,
  and J.~Apostolopoulos.
\newblock Connectboard: A remote collaboration system that supports gaze-aware
  interaction and sharing.
\newblock In {\em 2009 IEEE International Workshop on Multimedia Signal
  Processing}, pp. 1--6, 2009.

\bibitem{Tang2007}
A.~Tang, C.~Neustaedter, and S.~Greenberg.
\newblock Videoarms: Embodiments for mixed presence groupware.
\newblock In N.~Bryan-Kinns, A.~Blanford, P.~Curzon, and L.~Nigay, eds., {\em
  People and Computers XX --- Engage}, pp. 85--102. Springer London, 2007.

\bibitem{Tang91}
J.~C. Tang and S.~Minneman.
\newblock Videowhiteboard: Video shadows to support remote collaboration.
\newblock In {\em CHI}, p. 315–322. ACM, 1991.

\bibitem{tang2021towards}
X.~Tang, T.~Wang, and C.-W. Fu.
\newblock Towards accurate alignment in real-time 3d hand-mesh reconstruction.
\newblock In {\em ICCV}, pp. 11698--11707, 2021.

\bibitem{Towles2002}
H.~Towles, W.~chao Chen, R.~Yang, S.~uok Kum, H.~F.~N. Kelshikar, J.~Mulligan,
  K.~Daniilidis, H.~Fuchs, C.~C. Hill, N.~K.~J. Mulligan, L.~Holden,
  B.~Zeleznik, A.~Sadagic, and J.~Lanier.
\newblock 3d tele-collaboration over internet2.
\newblock In {\em International Workshop on Immersive Telepresence}, 2002.

\bibitem{Erp2015}
J.~B.~F. van Erp and A.~Toet.
\newblock Social touch in human–computer interaction.
\newblock {\em Frontiers in Digital Humanities}, 2, 2015.

\bibitem{wallace1981merging}
B.~A. Wallace.
\newblock Merging and transformation of raster images for cartoon animation.
\newblock In {\em SIGGRAPH}, pp. 253--262, 1981.

\bibitem{wang2020rgb2hands}
J.~Wang, F.~Mueller, F.~Bernard, S.~Sorli, O.~Sotnychenko, N.~Qian, M.~A.
  Otaduy, D.~Casas, and C.~Theobalt.
\newblock Rgb2hands: real-time tracking of 3d hand interactions from monocular
  rgb video.
\newblock {\em TOG}, 39(6):1--16, 2020.

\bibitem{wen2000}
W.-C. Wen, H.~Towles, L.~Nyland, G.~Welch, and H.~Fuchs.
\newblock Toward a compelling sensation of telepresence: demonstrating a portal
  to a distant (static) office.
\newblock In {\em IEEE Conference on Visualization}, pp. 327--333, 2000.

\bibitem{Wood2016}
E.~Wood, J.~Taylor, J.~Fogarty, A.~Fitzgibbon, and J.~Shotton.
\newblock Shadowhands: High-fidelity remote hand gesture visualization using a
  hand tracker.
\newblock In {\em Proceedings of the 2016 ACM International Conference on
  Interactive Surfaces and Spaces}, p. 77–84. ACM, 2016.

\bibitem{MRTouch2018}
R.~Xiao, J.~Schwarz, N.~Throm, A.~D. Wilson, and H.~Benko.
\newblock Mrtouch: Adding touch input to head-mounted mixed reality.
\newblock {\em IEEE TVCG}, 24(4):1653–1660, apr 2018.

\bibitem{yao2018mvsnet}
Y.~Yao, Z.~Luo, S.~Li, T.~Fang, and L.~Quan.
\newblock Mvsnet: Depth inference for unstructured multi-view stereo.
\newblock In {\em ECCV}, pp. 767--783, 2018.

\bibitem{Zhang2013}
C.~Zhang, Q.~Cai, P.~A. Chou, Z.~Zhang, and R.~Martin-Brualla.
\newblock Viewport: A distributed, immersive teleconferencing system with
  infrared dot pattern.
\newblock {\em IEEE MultiMedia}, 20(1):17--27, 2013.

\bibitem{Virtualcube2022}
Y.~Zhang, J.~Yang, Z.~Liu, R.~Wang, G.~Chen, X.~Tong, and B.~Guo.
\newblock Virtualcube: An immersive 3d video communication system.
\newblock {\em IEEE TVCG}, 28(5):2146--2156, 2022.

\bibitem{zhang2000flexible}
Z.~Zhang.
\newblock A flexible new technique for camera calibration.
\newblock {\em IEEE TPAMI}, 22(11):1330--1334, 2000.

\bibitem{Zillner2014}
J.~Zillner, C.~Rhemann, S.~Izadi, and M.~Haller.
\newblock 3d-board: A whole-body remote collaborative whiteboard.
\newblock In {\em UIST}, p. 471–479. ACM, 2014.

\bibitem{zollhofer2014real}
M.~Zollh{\"o}fer, M.~Nie{\ss}ner, S.~Izadi, C.~Rehmann, C.~Zach, M.~Fisher,
  C.~Wu, A.~Fitzgibbon, C.~Loop, C.~Theobalt, et~al.
\newblock Real-time non-rigid reconstruction using an rgb-d camera.
\newblock {\em ACM TOG}, 33(4):1--12, 2014.

\end{thebibliography}
\end{document}